\documentclass[sigconf]{acmart}

\usepackage{graphicx}
\usepackage{amsmath}
\usepackage{booktabs}
\usepackage{array}
\usepackage{multirow}
\usepackage{indentfirst}
\usepackage{subcaption}
\usepackage{caption}
\usepackage{float}
\usepackage{placeins}
\usepackage{xcolor}
\usepackage{hyperref}
\usepackage{afterpage}
\usepackage{lipsum}
\usepackage{cuted}

\hypersetup{
    colorlinks=true,
    linkcolor=blue,
    citecolor=blue,
    filecolor=blue,
    urlcolor=blue,
}

\AtBeginDocument{%
  }

\setcopyright{acmlicensed}
\copyrightyear{2024}
\acmYear{2024}
\acmDOI{XXXXXXX.XXXXXXX}

\acmConference[MM Asia '24]{ACM Multimedia Asia 2024}{December 03--06, 2024}{Auckland, New Zealand}
\acmISBN{978-1-4503-XXXX-X/24/12}

\author{Liang Feng}
\email{2210485001@email.szu.edu.cn}
\affiliation{%
  \institution{Shenzhen University}
  \city{Shenzhen}
  \country{China}
}

\author{Ming Xu}
\authornote{Corresponding author.}
\email{xuming@szu.edu.cn}
\affiliation{%
  \institution{Shenzhen University}
  \city{Shenzhen}
  \country{China}
}

\author{Lihua Wen}
\authornote{Both authors contributed equally to this research.}
\email{2210485002@email.szu.edu.cn}
\affiliation{%
  \institution{Shenzhen University}
  \city{Shenzhen}
  \country{China}
}

\author{Zhixuan Shen}
\authornotemark[2]
\email{2310485003@email.szu.edu.cn}
\affiliation{%
  \institution{Shenzhen University}
  \city{Shenzhen}
  \country{China}
}

\begin{document}

\title{GatedUniPose: A Novel Approach for Pose Estimation Combining UniRepLKNet and Gated Convolution}

\begin{abstract}
Pose estimation is a crucial task in computer vision, with wide applications in autonomous driving, human motion capture, and virtual reality. However, existing methods still face challenges in achieving high accuracy, particularly in complex scenes. This paper proposes a novel pose estimation method, GatedUniPose, which combines UniRepLKNet and Gated Convolution and introduces the GLACE module for embedding. Additionally, we enhance the feature map concatenation method in the head layer by using DySample upsampling. Compared to existing methods, GatedUniPose excels in handling complex scenes and occlusion challenges. Experimental results on the COCO, MPII, and CrowdPose datasets demonstrate that GatedUniPose achieves significant performance improvements with a relatively small number of parameters, yielding better or comparable results to models with similar or larger parameter sizes.
\end{abstract}



\maketitle

\begin{figure}[t]
    \centering
    \includegraphics[width=\linewidth]{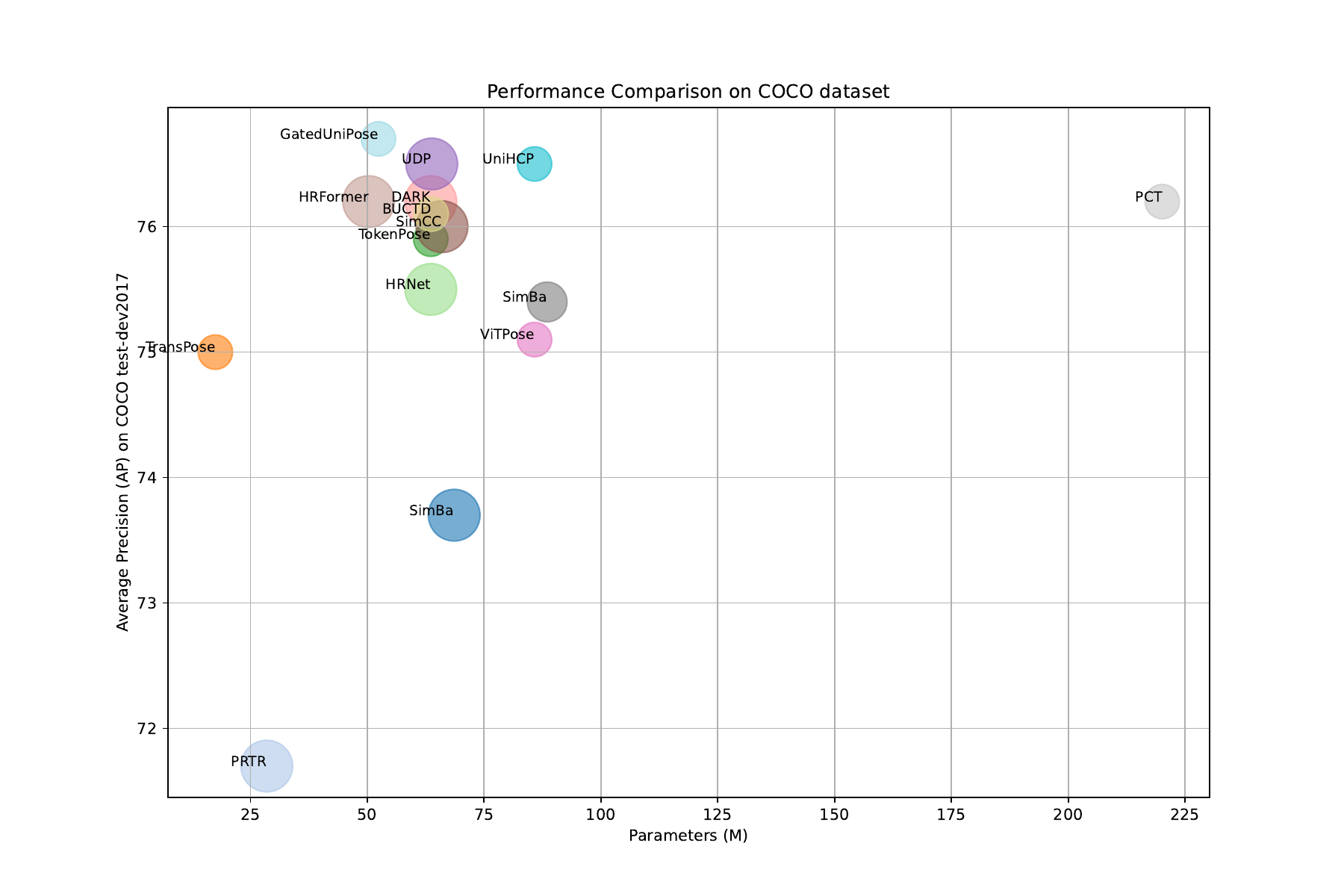}
    \caption{The comparison of GatedUniPose and advanced methods on the COCO test-dev2017 set regarding model size and precision. The size of each bubble represents the input size of the model.}
    \label{fig:model_architecture}
\end{figure}

\section{Introduction}
Human pose estimation is a fundamental task in computer vision that involves estimating body joint positions from images. This task is crucial for various applications, including autonomous driving, human motion capture, and virtual reality \cite{cao2017realtime,Chen2020.12.04.405159}. Recent advancements in human pose estimation have significantly improved accuracy on public datasets through innovations in network architectures, training methodologies, and fusion strategies. For instance, methods such as OpenPose \cite{cao2017realtime} and AlphaTracker \cite{Chen2020.12.04.405159} have set new benchmarks in the field.

Despite these advancements, existing methods still face significant challenges in complex scenes, such as occlusion and lighting variations \cite{li2019crowdpose,zhang2019pose2seg}, which limit their effectiveness in real-world applications. Current 2D and 3D pose estimators typically use coordinate vectors or heatmap embeddings to represent poses. However, these representations often fail to model the dependencies between joints, leading to unrealistic pose estimations in challenging scenarios.

Humans, on the other hand, can effortlessly predict complete poses using contextual information, even when only partial joints are visible. This observation suggests that context is crucial for accurate pose estimation. Motivated by these challenges, our research aims to learn the dependencies between joints rather than relying on manually designed rules, inspired by the human ability to use contextual information for pose prediction.

This paper proposes a novel pose estimation method, GatedUniPose, which combines UniRepLKNet \cite{ding2024unireplknet} and Gated Convolution \cite{dauphin2017language} and introduces the GLACE module \cite{zhang2017alignedreid} for embedding. Additionally, we enhance the feature map concatenation method in the head layer by using DySample upsampling \cite{liu2023learning}.

Our contributions are as follows: introducing Gated Convolution \cite{dauphin2017language} in UniRepLKNet \cite{ding2024unireplknet} to improve feature extraction; using the GLACE module \cite{zhang2017alignedreid} for embedding, thereby enhancing accuracy; improving the feature map concatenation method in the head layer by using DySample \cite{liu2023learning} upsampling; addressing the occlusion handling limitation in PCT \cite{geng2023human}, the specific task performance limitation in UniHCP \cite{ci2023unihcp}, and the generalization capability issue in complex scenes in BUCTD \cite{zhou2023rethinking}; and achieving comparable or even superior accuracy with significantly fewer parameters compared to state-of-the-art methods.

Our experimental results demonstrate that GatedUniPose achieves state-of-the-art performance in general scenarios and exhibits robustness in occluded conditions. The code and models will be made publicly available to facilitate further research.

\section{Related Work}
\subsection{Pose Estimation Methods and Benchmarks}
Pose estimation methods can be broadly categorized into top-down and bottom-up approaches. Top-down methods \cite{fang2017rmpe,Chen2020.12.04.405159,wang2020deep,yang2021transpose,li2021tokenpose,mao2103tfpose,xu2022vitpose} first detect individual instances using an object detector \cite{ren2015faster,redmon2018yolov3,he2017mask,jocher2021ultralytics,liu2021swin} and then estimate the pose within each detected bounding box. Bottom-up methods \cite{cao2017realtime,newell2017associative,insafutdinov2017arttrack,cheng2020higherhrnet,geng2021bottom,lauer2022multi} first detect all body parts in the image and then group them into individual instances. Recently, transformer-based methods \cite{yang2021transpose,li2021tokenpose,mao2103tfpose,xu2022vitpose} have shown promising results in pose estimation tasks.

Classic benchmarks for human pose estimation, such as COCO \cite{lin2014microsoft} and MPII \cite{andriluka20142d}, contain few occlusions \cite{khirodkar2021multi}, which are not representative of many real-world scenarios. Recently, new benchmarks with more crowded scenes, such as CrowdPose \cite{li2019crowdpose} and OCHuman \cite{zhang2019pose2seg}, have emerged to address this issue. Multi-animal pose estimation benchmarks \cite{mathis2020deep,lauer2022multi} also share some challenges with human benchmarks, such as heavy overlap and similar appearances within a species.

\subsection{Pose Estimation in Crowded Scenes}
Crowded scenes pose significant challenges for pose estimation due to occlusions and the dense grouping of individuals. Several works \cite{li2019crowdpose,zhang2019pose2seg,qiu2020peeking,khirodkar2021multi} have focused on addressing these challenges. Khirodkar et al. \cite{khirodkar2021multi} proposed a hybrid top-down approach called MIPNet, which predicts multiple people within a given bounding box. CID \cite{wang2022contextual} and CenterGroup \cite{braso2021center} use attention mechanisms to distinguish between individuals and link body parts. PETR \cite{shi2022end} deploys separate transformer-based decoders for individuals and keypoints.

\subsection{Combining Top-Down and Bottom-Up Models}
Several works have explored combining top-down and bottom-up approaches for pose estimation. Hu and Ramanan \cite{hu2016bottom} proposed a bidirectional architecture that incorporates top-down feedback with a bottom-up architecture. Tang et al. \cite{tang2018deeply} introduced a hierarchical, compositional model with both bottom-up and top-down stages. Cai et al. \cite{cai2019exploiting} and Li et al. \cite{li2019multi} proposed methods that use bottom-up approaches to estimate joints and top-down information to group them.

\subsection{Gated Convolution and Advanced Embedding Techniques}
Gated Convolution (GConv) \cite{dauphin2017language} has shown promising results in various computer vision tasks by introducing a gating mechanism to control information flow and enhance feature extraction capabilities. The GLACE module \cite{zhang2017alignedreid} has demonstrated significant feature extraction effects in other tasks but has not yet been applied to pose estimation. UniRepLKNet \cite{ding2024unireplknet}, an efficient feature extraction network, has been widely used in various computer vision tasks but has not yet been fully explored in pose estimation.

Our work, GatedUniPose, addresses the challenges in pose estimation by combining UniRepLKNet \cite{ding2024unireplknet}, Gated Convolution \cite{dauphin2017language}, and the GLACE module \cite{zhang2017alignedreid}. We also enhance the feature map concatenation method in the head layer by using DySample upsampling \cite{liu2023learning}. Compared to existing methods, GatedUniPose excels in handling complex scenes and occlusion challenges while achieving a better balance between parameter count and accuracy. Our work provides an important reference for combining gated convolution and advanced embedding techniques to advance the field of pose estimation.

\section{Methodology}

\subsection{Overall Architecture}

The overall architecture of GatedUniPose is illustrated in Figure \ref{fig:model_architecture}. The input image size is [3, 256, 192]. First, the image is downsampled to [768, 128, 96] using the GLACE \cite{zhang2017alignedreid} and then input into the improved GatedUniPose backbone. The first downsampling layer of each stage in the backbone is implemented using the GLACE, and Gated Convolution \cite{dauphin2017language} is integrated into the blocks. The final feature maps are upsampled to the same size using the DySample module \cite{liu2023learning}, concatenated in the head layer, and finally fed into the decoder layer.

\begin{figure*}[t]
    \centering
    \includegraphics[width=\linewidth]{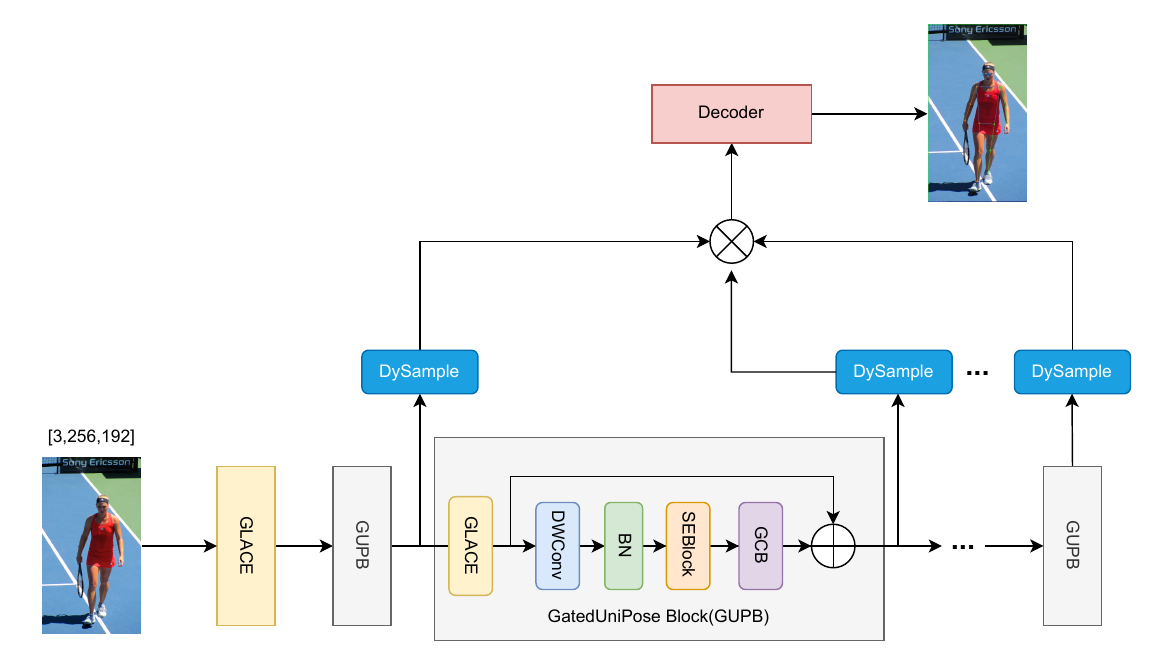}
    \caption{The overall architecture of GatedUniPose.}
    \label{fig:model_architecture}
\end{figure*}

\begin{figure*}[t]
    \centering
    \begin{minipage}[t]{0.48\textwidth}
        \centering
        \includegraphics[width=\linewidth]{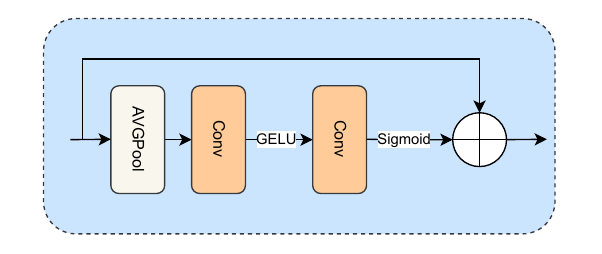}
        \caption*{(a) The overall architecture of SEBlock.}
        \label{fig:seblock}
    \end{minipage}
    \hfill
    \begin{minipage}[t]{0.48\textwidth}
        \centering
        \includegraphics[width=\linewidth]{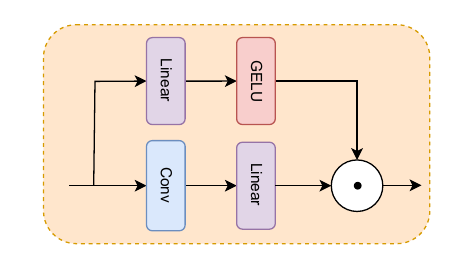}
        \caption*{(b) The overall architecture of GCB.}
        \label{fig:gcb}
    \end{minipage}
\end{figure*}

\subsection{GatedUniPose Backbone}

The backbone of GatedUniPose is based on UniRepLKNet \cite{ding2024unireplknet} with the following enhancements:

\begin{itemize}
    \item \textbf{GLACE Integration}: The GLACE \cite{zhang2017alignedreid} is utilized in the first downsampling layer of each stage to enhance initial feature extraction capabilities. The GLACE downsamples the input image to [768, 128, 96], and its output serves as the input for subsequent stages.
    
    \item \textbf{Gated Convolution Integration}: Gated Convolution (GConv) \cite{dauphin2017language} is integrated into each UniRepLKNetBlock to enhance feature extraction capabilities.
    \[
    \text{GConv}(x) = \sigma(W_g * x) \odot (W_v * x)
    \]
    where \(\sigma\) is the Sigmoid activation function, \(\odot\) denotes element-wise multiplication, and \(W_g\) and \(W_v\) are the weights of the gated convolution and regular convolution, respectively.
    
    \item \textbf{GatedUniPoseBlock}: The structure of each GatedUniPoseBlock is as follows:
    \[
    \begin{aligned}
    &\text{If } \text{kernel\_size} \geq 7: \quad \text{Dilated Reparam Block} \\
    &\text{Else:} \quad \text{Depthwise Convolution}
    \end{aligned}
    \]
\end{itemize}

\subsection{Head Layer and DySample Upsampling}

In the head layer, we utilize the DySample module \cite{liu2023learning} to upsample the feature maps output by each stage of the backbone to the same size and concatenate them.
\[
F_{\text{up}} = \text{DySample}(F_{\text{stage}})
\]
where \(F_{\text{stage}}\) is the output feature map of each stage, and \(F_{\text{up}}\) is the upsampled feature map. The concatenated feature map serves as the input for the final pose estimation.

\begin{table*}
    \centering
    \label{tab:performance_comparison}
    \renewcommand{\arraystretch}{1.5}
    \resizebox{\textwidth}{!}{
    \begin{tabular}{|>{\centering\arraybackslash}p{2.5cm}|>{\centering\arraybackslash}p{3cm}>{\centering\arraybackslash}p{2.5cm}>{\centering\arraybackslash}p{2cm}|>{\centering\arraybackslash}p{1.5cm}>{\centering\arraybackslash}p{1.5cm}>{\centering\arraybackslash}p{1.5cm}|>{\centering\arraybackslash}p{1.5cm}>{\centering\arraybackslash}p{1.5cm}>{\centering\arraybackslash}p{1.5cm}|}
        \hline
        \multirow{2}{*}{Method} & \multirow{2}{*}{Backbone} & \multirow{2}{*}{Input size} & \multirow{2}{*}{Parms (M)} & \multicolumn{3}{c|}{COCO test-dev2017 $\uparrow$} & \multicolumn{3}{c|}{COCO val2017 $\uparrow$} \\
        \cline{5-10}
        & & & & AP & AP$^{50}$ & AP$^{75}$ & AP & AP$^{50}$ & AP$^{75}$ \\
        \hline
        SimBa. \cite{xiao2018simple} & ResNet-152 & 384 $\times$ 288 & 68.6 & 73.7 & 91.9 & 81.1 & 74.3 & 89.6 & 81.1 \\
        PRTR \cite{li2021pose} & HRNet-W32 & 384 $\times$ 288 & 28.5 & 71.7 & 90.6 & 79.6 & 73.1 & 89.4 & 79.8 \\
        TransPose \cite{yang2021transpose} & HRNet-W48 & 256 $\times$ 192 & 17.5 & 75.0 & 92.2 & 82.3 & 75.8 & 90.1 & 82.1 \\
        TokenPose \cite{li2021tokenpose} & HRNet-W48 & 256 $\times$ 192 & 63.6 & 75.9 & 92.3 & 83.4 & 75.8 & 90.3 & 82.5 \\
        HRNet \cite{sun2019deep} & HRNet-W48 & 384 $\times$ 288 & 63.6 & 75.5 & 92.7 & 83.3 & 76.3 & 90.8 & 82.9 \\
        DARK \cite{zhang2020distribution} & HRNet-W48 & 384 $\times$ 288 & 63.6 & 76.2 & 92.5 & 83.6 & 76.8 & 90.6 & 83.2 \\
        UDP \cite{huang2020devil} & HRNet-W48 & 384 $\times$ 288 & 63.8 & 76.5 & 92.7 & 84.0 & 77.8 & 92.0 & 84.3 \\
        SimCC \cite{li20212d} & HRNet-W48 & 384 $\times$ 288 & 66.0 & 76.0 & 92.4 & 83.5 & 76.9 & 90.9 & 83.2 \\
        HRFormer \cite{yuan2110hrformer} & HRFormer-B & 384 $\times$ 288 & 50.3 & 76.2 & 92.7 & 83.8 & 77.2 & 91.0 & 83.6 \\
        ViTPose \cite{xu2022vitpose} & ViT-Base & 256 $\times$ 192 & 85.8 & 75.1 & 92.5 & 83.1 & 75.8 & 90.7 & 83.2 \\
        SimBa. \cite{xiao2018simple} & Swin-Base & 256 $\times$ 256 & 88.5 & 75.4 & 93.0 & 84.1 & 76.6 & 91.4 & 84.3 \\
        PCT \cite{geng2023human} & Swin-Base & 256 $\times$ 192 & 220.1 & 76.2 & 92.1 & 84.5 & 77.2 & 91.2 & 84.3 \\
        BUCTD \cite{zhou2023rethinking} & HRNet-W48 & 256 $\times$ 192 & 63.7 & 76.1 & 92.5 & 84.2 & 76.8 & 91.1 & 84.5 \\
        UniHCP \cite{ci2023unihcp} & ViT-Base & 256 $\times$ 192 & 85.8 & 76.5 & 92.5 & 84.2 & 76.8 & 91.1 & 84.5 \\
        \hline
        Our approach & UniRepLKNet-B & 256 $\times$ 192 & 52.4 & \textbf{76.7} & 90.7 & 83.4 & \textbf{77.4} & 90.7 & 84.2 \\
        \hline
    \end{tabular}
    }
    \caption{Performance comparison on COCO dataset}
\end{table*}

\subsection{Decoder and Loss Function}

We utilize a classic decoder layer for the final pose estimation. The formula for calculating the output heatmap of the decoder is as follows:
\[
K = \text{Conv}_{1 \times 1}(\text{Deconv}(\text{Deconv}(F_{\text{out}})))
\]
where \(F_{\text{out}}\) is the concatenated feature map from the head layer, and \(K\) is the output heatmap.

The loss function includes output distillation loss and token-based knowledge distillation loss:
\[
L_{\text{od}}^{t \rightarrow s} = \text{MSE}(K_s, K_t)
\]
\[
t^* = \arg \min_t (\text{MSE}(T(t; X), K_{\text{gt}}))
\]

With these enhancements, our GatedUniPose model excels in handling complex scenes and occlusion challenges while achieving a good balance between the number of parameters and accuracy.

\section{Experiments}

To evaluate GatedUniPose, we performed comprehensive experiments on several benchmarks. We tested our approach on the most important benchmarks for pose estimation, including COCO \cite{lin2014microsoft}, CrowdPose \cite{li2019crowdpose}, and MPII \cite{andriluka20142d}. We also carried out several ablations to test the design choices.

\subsection{COCO Benchmark}

\textbf{Dataset:} The COCO dataset \cite{lin2014microsoft} contains 57K images with 150K persons in the train set, 5K images with 6.3K persons in the val set, and 20K images in the test-dev set. We used the train set for training and the val set for validation. We report standard metrics including AP, AP\textsubscript{50}, and AP\textsubscript{75}.

\textbf{Results:} We compared GatedUniPose with several state-of-the-art methods. The results are summarized in Table 1. GatedUniPose consistently outperforms these methods in terms of AP on the COCO dataset. Notably, GatedUniPose achieves an AP of 76.7\% on COCO test-dev2017 and 77.4\% on COCO val2017 with only 52.4M parameters, significantly outperforming other advanced methods. This demonstrates that GatedUniPose strikes a good balance between accuracy and efficiency.

\subsection{MPII Benchmark}

\textbf{Dataset:} The MPII dataset \cite{andriluka20142d} contains around 25K images with over 40K annotated poses. We used the standard train/test split provided by the dataset. We report the PCKh (Percentage of Correct Keypoints with head-normalized distance) metric.

\textbf{Results:} We assessed the performance of GatedUniPose on the MPII dataset and compared it with several state-of-the-art methods. The results are presented in Table 2. GatedUniPose demonstrates superior performance in terms of PCKh, highlighting its robustness across various scenarios.

\subsection{CrowdPose Benchmark}

\textbf{Dataset:} The CrowdPose dataset \cite{li2019crowdpose} contains 12K labeled images in the trainval set with 43.4K labeled people (each with 14 keypoints), and 8K images in the test set with 29K labeled people. Following other studies, we used the trainval set for training and the test set for evaluation. We report standard metrics AP.

\textbf{Results:} We trained GatedUniPose on the CrowdPose dataset and compared it with several advanced methods. The results, shown in Table 3, indicate that while GatedUniPose performs competitively, it does not achieve the highest performance on CrowdPose. However, GatedUniPose still demonstrates strong performance on other datasets such as COCO and MPII, indicating its overall robustness and effectiveness.

\begin{table*}
    \centering
    \renewcommand{\arraystretch}{1.5}
    \setlength{\tabcolsep}{10pt} 
    \resizebox{\textwidth}{!}{
    \begin{tabular}{|p{3cm}|c|c|c|c|c|c|c|c|}
        \hline
        Method & Hea. & Sho. & Elb. & Wri. & Hip. & Kne. & Ank. & Mean \\
        \hline
        SimBa. \cite{xiao2018simple} & 97.0 & 95.6 & 90.0 & 86.2 & 89.7 & 86.9 & 82.9 & 90.2 \\
        PRTR \cite{li2021pose} & 97.3 & 96.0 & 90.6 & 84.5 & 89.7 & 85.5 & 79.0 & 89.5 \\
        HRNet \cite{sun2019deep} & 97.1 & 95.9 & 90.3 & 86.4 & 89.7 & 88.3 & 83.3 & 90.3 \\
        DARK \cite{zhang2020distribution} & 97.2 & 95.9 & 91.2 & 86.7 & 89.7 & 86.7 & 84.7 & 90.3 \\
        TokenPose \cite{li2021tokenpose}  & 97.1 & 95.9 & 90.4 & 85.6 & 89.5 & 85.8 & 81.8 & 89.4 \\
        SimCC \cite{li20212d} & 97.2 & 96.0 & 90.4 & 85.6 & 89.5 & 85.8 & 81.8 & 90.0 \\
        \hline
        Our approach & 97.3 & 96.0 & 90.8 & 86.7 & 89.4 & 86.6 & 82.3 & 90.2 \\
        \hline
    \end{tabular}
    }
    \caption{Performance comparison on MPII dataset}
    \label{tab:keypoint_detection}
\end{table*}

\begin{table*}
    \centering
    \begin{minipage}{0.48\textwidth}
        \centering
        \label{tab:occluded_pose_estimation}
        \renewcommand{\arraystretch}{1.5}
        \setlength{\tabcolsep}{10pt} 
        \resizebox{\textwidth}{!}{
        \begin{tabular}{|p{5cm}|c|}
            \hline
            Method & mAP \\
            \hline
            SCIO \cite{kan2022self} & 71.5 \\
            BUCTD \cite{zhou2023rethinking} & 72.9 \\
            KAPAO-L \cite{mcnally2022rethinking} & 68.9 \\
            MIPNet \cite{khirodkar2021multi} & 70.0 \\
            OpenPifPaf \cite{kreiss2021openpifpaf} & 70.5 \\
            \hline
            Our approach & 70.8 \\
            \hline
        \end{tabular}
        }
        \caption{Performance comparison on CrowdPose dataset}
    \end{minipage}
    \hfill
    \begin{minipage}{0.48\textwidth}
        \centering
        \label{tab:ablation_study}
        \renewcommand{\arraystretch}{1.5}
        \setlength{\tabcolsep}{10pt} 
        \resizebox{\textwidth}{!}{
        \begin{tabular}{|c|c|c|c|c|}
            \hline
            UniRepLKNet & GatedUniPose & GLACE & DySample & COCO/AP \\
            \hline
            $\checkmark$  & & & & 75.7\\
            \hline
             & $\checkmark$ & & & 76.1\\
             \hline
            $\checkmark$  & & $\checkmark$&  $\checkmark$& 76.0\\
            \hline
             & $\checkmark$ & &  $\checkmark$& 76.3\\
            \hline
            & $\checkmark$ & $\checkmark$& & 76.5\\
            \hline
            & $\checkmark$ & $\checkmark$&  $\checkmark$& 76.7\\
            \hline
        \end{tabular}
        }
        \caption{Ablation Study on COCO Dataset}
    \end{minipage}
\end{table*}

\subsection{Ablation Studies}

\textbf{Setup:} To assess the contribution of each component in our model, we conducted a series of ablation studies. Specifically, we evaluated the effects of using only Gated Convolution \cite{dauphin2017language}, only the GLACE module \cite{zhang2017alignedreid}, and only the DySample \cite{liu2023learning} upsampling module. The experiments were performed on the COCO dataset, and the results are reported in terms of Average Precision (AP).

\textbf{Results:} The results of the ablation studies are summarized in Table 4. Each component of the model contributes significantly to the overall performance. The inclusion of Gated Convolution\cite{dauphin2017language} and the GLACE module  \cite{zhang2017alignedreid} enhances the feature extraction capabilities, while the DySample upsampling \cite{liu2023learning} module improves feature map fusion. These enhancements collectively lead to a notable improvement in pose estimation accuracy, demonstrating the effectiveness of each individual component as well as their synergistic integration.

\subsection{Results Analysis}

The experimental results indicate that GatedUniPose excels in complex scenarios, particularly those involving occlusion and lighting variations. Compared to other advanced methods, our approach demonstrates superior efficiency in multi-task processing, effectively reducing resource consumption. Additionally, GatedUniPose shows enhanced performance in handling occlusion issues and specific tasks. Notably, GatedUniPose achieves comparable or even superior performance with a relatively smaller parameter count, highlighting its efficiency and effectiveness.

\section{Conclusion}

In this work, we introduce GatedUniPose, a novel pose estimation method that leverages the strengths of UniRepLKNet \cite{ding2024unireplknet} and Gated Convolution \cite{dauphin2017language}, while incorporating the GLACE module \cite{zhang2017alignedreid} for enhanced embedding. Additionally, we improve feature map concatenation in the head layer using DySample upsampling. Our extensive experiments on the COCO \cite{lin2014microsoft} and MPII \cite{andriluka20142d} datasets demonstrate that GatedUniPose achieves significant performance improvements over existing methods such as PCT \cite{geng2023human}, UniHCP \cite{ci2023unihcp}, and BUCTD \cite{zhou2023rethinking}. Notably, GatedUniPose excels in handling complex scenes and occlusion challenges, while maintaining a relatively smaller parameter count, underscoring its efficiency and effectiveness.

\bibliographystyle{ACM-Reference-Format}
\bibliography{references}

\end{document}